\documentclass[letterpaper, 10 pt, conference]{ieeeconf}  

\IEEEoverridecommandlockouts                              

\makeatletter
\let\NAT@parse\undefined
\makeatother

\usepackage{color}
\usepackage[table,dvipsnames]{xcolor}

\usepackage{graphics} 
\usepackage{mathptmx} 
\usepackage{times} 
\usepackage{amsmath} 
\usepackage{amssymb}  
\usepackage{caption}
\usepackage{graphicx}
\usepackage{booktabs}
\let\labelindent\relax
\usepackage{nicematrix,enumitem}
\usepackage{makecell}
\usepackage{comment}
\usepackage{wrapfig}
\usepackage{balance}
\usepackage{tablefootnote}
\usepackage[square, comma, numbers,sort&compress]{natbib}
\usepackage{dsfont}
\usepackage{multirow}
\usepackage{pifont}

\usepackage[pagebackref=true,breaklinks=true,colorlinks,bookmarks=false]{hyperref}

\hypersetup{
linkcolor=BrickRed
,citecolor=Green
,filecolor=Mulberry
,urlcolor=NavyBlue
,menucolor=BrickRed
,runcolor=Mulberry
,linkbordercolor=BrickRed
,citebordercolor=Green
,filebordercolor=Mulberry
,urlbordercolor=NavyBlue
,menubordercolor=BrickRed
,runbordercolor=Mulberry
}

\title{\LARGE \bf
NYC-Event-VPR: A Large-Scale High-Resolution Event-Based Visual Place Recognition Dataset in Dense Urban Environments
}

\author{Taiyi Pan, Junyang He, Chao Chen$^{*}$, Yiming Li$^{*}$, and Chen Feng\textsuperscript{\ding{41}}
\thanks{$^{*}$ Equal contributions.}
\thanks{\ding{41} The authors are with New York University, Brooklyn, NY 11201, USA. Chen Feng is the corresponding author ({\tt\small \href{mailto:cfeng}{cfeng@nyu.edu}}). This work is supported by NSF Grant 2238968, and in part through the NYU IT High Performance Computing resources, services, and staff expertise.}
\\
{\tt\small\url{https://ai4ce.github.io/NYC-Event-VPR}
}}

\IEEEaftertitletext{
\begin{center}
    \vspace{-7mm}
    \centering
    \captionsetup{type=figure}
    \includegraphics[width=\linewidth,height=6\baselineskip]
    {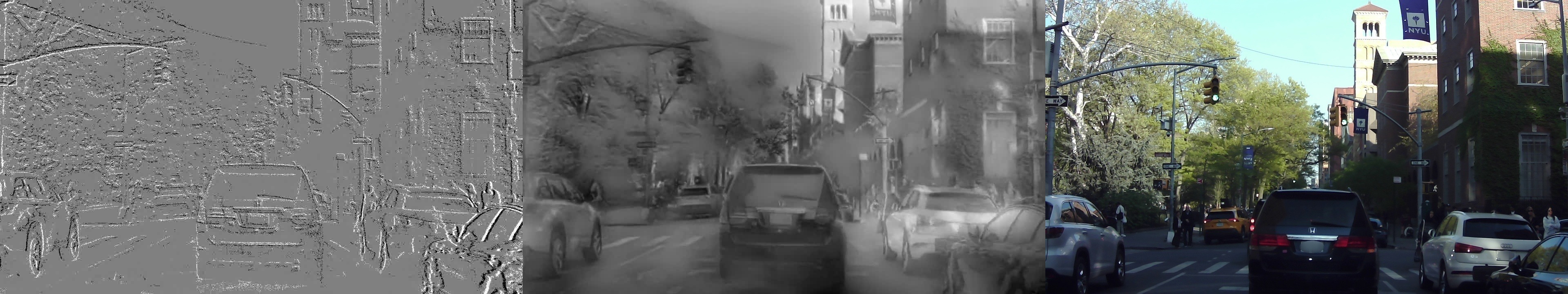}
  \captionof{figure}{ Downtown Manhattan in the event frame, reconstructed frame, and RGB frame.}
  \label{figure:showcase}
\end{center}%
}

\begin{document}

\maketitle
\thispagestyle{empty}
\pagestyle{empty}


\begin{abstract}

Visual place recognition (VPR) enables autonomous robots to identify previously visited locations, which contributes to tasks like simultaneous localization and mapping (SLAM). VPR faces challenges such as accurate image neighbor retrieval and appearance change in scenery. Event cameras, also known as dynamic vision sensors, are a new sensor modality for VPR and offer a promising solution to the challenges with their unique attributes: high temporal resolution (1MHz clock), ultra-low latency (in \textmu s), and high dynamic range ($>$120dB). These attributes make event cameras less susceptible to motion blur and more robust in variable lighting conditions, making them suitable for addressing VPR challenges. However, the scarcity of event-based VPR datasets, partly due to the novelty and cost of event cameras, hampers their adoption. To fill this data gap, our paper introduces the NYC-Event-VPR dataset to the robotics and computer vision communities, featuring the Prophesee IMX636 HD event sensor (1280x720 resolution), combined with RGB camera and GPS module. It encompasses over 13 hours of geotagged event data, spanning 260 kilometers across New York City, covering diverse lighting and weather conditions, day/night scenarios, and multiple visits to various locations. Furthermore, our paper employs three frameworks to conduct generalization performance assessments, promoting innovation in event-based VPR and its integration into robotics applications.

\end{abstract}


\section{Introduction}

\begin{figure}[t]
    \centering
    \includegraphics[width=0.95\columnwidth]{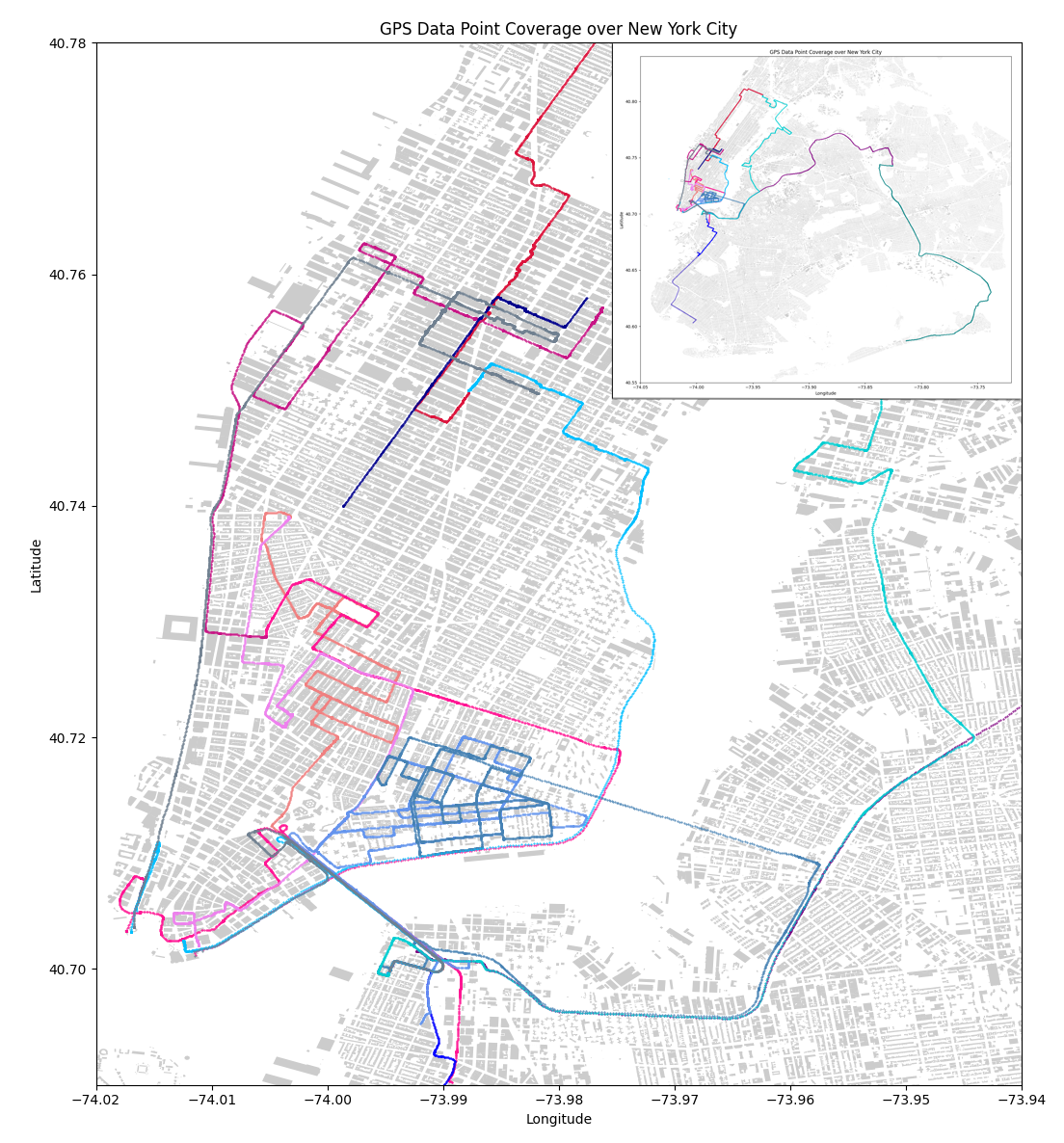}
    \caption{NYC-Event-VPR covers NYC, focusing on Chinatown area in Manhattan with overlapping traversal. Each color represents a distinct traversal.}
    \vspace{-6mm}
    \label{fig:coverage}
\end{figure}

Visual place recognition (VPR) aims to enable autonomous robots to determine whether they have previously visited a particular location. It plays a pivotal role in tasks like providing loop closure in visual Simultaneous Localization and Mapping (SLAM) systems \cite{lowry2015visual}. Typically, RGB images are used for visual input to conduct VPR tasks \cite{arandjelovic2016netvlad}. Beyond RGB image input, various other sensor modalities for VPR exist, including using radio detection and ranging (radar) \cite{cai2022autoplace} and using light detection and ranging (LiDAR)~\cite{guo2019local, kim2020mulran}. In general, VPR poses significant challenges including the need for accurate neighbor retrieval and the appearance change of scenery over time and under different lighting conditions \cite{zhang2021visual}. Event cameras represent another sensor modality in VPR task domain. They potentially offer a unique solution to the challenge of appearance change in VPR since they are less susceptible to motion blurs and are more invariant to illumination changes in visual scenery.

\begin{table*}[t!]
\caption{\textbf{Event-based VPR dataset attribute comparisons} (NYC-Event-VPR with highest resolution)}
\vspace{-3mm}
\label{table:dataset_comparison}
\begin{center}
\resizebox{\linewidth}{!}{
\begin{NiceTabular}{c c c c c c}
\toprule
\textbf{Dataset} \tabularnote{Other event-based VPR datasets (non-exhaustive): DDD17~\cite{binas2017ddd17}, MVSEC~\cite{zhu2018multivehicle}, Oxford RobotCar~\cite{maddern20171} (synthetic), CARLA~\cite{dosovitskiy2017carla} (synthetic)} & \textbf{Event Camera} & \textbf{Resolution (px)} & \textbf{Duration (hr)} & \textbf{Traversal distance (km)} & \textbf{Main locations} \\
\hline
Brisbane-Event-VPR~\cite{fischer2020event} &
    IniVation DAVIS346
 & 346x260 & 1.13 & 48 & suburbs of Brisbane, Australia \\
\hline
DDD20~\cite{hu2020ddd20} &
    IniVation DAVIS346
 & 346x260 & 51 & 4000 & southwestern USA \\
\hline
DSEC~\cite{gehrig2021dsec} &
    Prophesee PPS3MVCD
 & 640x480 & 0.887 & - & near Zurich, Switzerland\\
\hline
eTraM~\cite{Verma_2024_CVPR} &
    Prophesee EVK4 HD
 & 1280x720 & 10 & - & Tempe, USA\\
\hline
NYC-Event-VPR &
    Prophesee EVK4 HD
 & 1280x720 & 13.5 & 259.95 & New York City, USA \\
\bottomrule
\end{NiceTabular}
}
\end{center}
\vspace{-6mm}
\end{table*}

Event cameras, also known as dynamic vision sensors, offer several distinct advantages over frame-based cameras, including high temporal resolution (1MHz clock), ultra-low latency (in microseconds), high dynamic range (exceeding 120dB), and minimal power consumption when idle (10mW) \cite{gallego2020event}. The pixels in an event sensor operate asynchronously, responding independently to changes in illumination for each pixel without relying on an external clock for sampling. Raw event data is represented as a continuous sequence of 4-dimensional vectors $\begin{pmatrix} x & y & p & t \end{pmatrix}$ containing pixel coordinates x and y, polarity p (0 or 1), and a time stamp t in \textmu s resolution. The high temporal resolution and low latency aspect of event camera ensure minimal motion blur in event data, enhancing image neighbor retrieval. Furthermore, their robustness to appearance change in visual scenery, thanks to the high dynamic range, allows event cameras to perform well in challenging lighting conditions ~\cite{Shariff_event_review_2024}. These characteristics make event cameras a promising technology for addressing VPR challenges, as well as other autonomous vehicle tasks.

Despite these benefits, the literature contains only a handful of event-based VPR datasets, primarily situated in sparse outdoor settings and using event cameras with lower resolutions. The lack of available event-based VPR datasets could be attributed to the fact that event cameras are a novel type of visual sensors and that costs of such novel sensors are high. This means event cameras are yet to be mainstream-adopted and that literature on event-based vision could be greatly expanded. This paper aims to address this data gap by contributing to the robotics and computer vision communities a novel event-based VPR dataset designed to inspire innovation in event-based vision:

\begin{enumerate}
    \item We introduce a large-scale high-resolution event-based VPR dataset captured within the bustling urban landscape of New York City as shown in Fig. \ref{fig:coverage}. This dataset is obtained using the Prophesee EVK4 HD evaluation kit, incorporating the IMX636 HD event sensor from Prophesee and Sony \cite{finateu20205}, delivering event data at a resolution of 1280x720. Our dataset is comprised of more than 13 hours of footage, covering 260km (Tab. \ref{table:statistics}). Our data collection spans multiple months, capturing diverse seasons, weather conditions, and lighting scenarios, and included event, RGB, and GPS data.
    \item To ensure seamless integration with the frameworks ~\cite{zaffar2021vpr, Berton_CVPR_2022_benchmark, Berton_2023_EigenPlaces} we utilize for benchmark experiments, we develop a data processing pipeline. This pipeline facilitates the creation of tailored datasets from raw sensor readings that align with frameworks' dataset specifications, enabling us to conduct initial benchmark evaluations.

\end{enumerate}

\section{Related work}\label{sec:related}

\textbf{Classic VPR.} In the field of visual place recognition (VPR) research, a multitude of datasets have been made available to facilitate the training and evaluation of new VPR techniques \cite{zhang2021visual}. These datasets contribute significantly to advancing the field. Notable examples include the Nordland outdoor dataset introduced by Skrede et al. \cite{Skrede_2013} in 2013, which presented a significant challenge for VPR due to substantial perceptual aliasing issues. Arandjelovic et al. \cite{arandjelovic2016netvlad} made significant contributions in 2016 by introducing the Pittsburgh 250k and Tokyo 24/7 datasets, accompanied by the NetVLAD architecture. These twin datasets offered diverse viewpoints of the same locations, enhancing dataset diversity. In the same year, Ros et al. \cite{ros2016synthia} presented the SYNTHIA dataset, which was artificially generated using the Unity engine. This dataset proved versatile, catering to various task domains, including VPR. In 2021, Sheng et al. \cite{sheng2021nyu} introduced the NYU-VPR dataset, which featured repeated traversals of New York City from a moving vehicle over the course of a year, capturing both front views and side views as the vehicle navigated the city streets. More recently, in 2022, Ali-Bey et al. \cite{ali2022gsv} introduced GSV-Cities, a large-scale dataset specifically designed for VPR. The dataset comprises of images organized into more than 62k distinct places, with a minimum of four images representing each place.

\label{other_event_vpr}
\textbf{Event-based VPR.} As we delve into the realm of event-based VPR literature, datasets become scarcer. Several datasets similar to ours are described in Tab. \ref{table:dataset_comparison} for reference. For instance, Fischer et al. \cite{fischer2020event} introduced the Brisbane-Event-VPR dataset, based in rural Brisbane, Australia. It includes 48km of repeated traversals with a total of 1.13 hours of geotagged footage collected over several days, encompassing both daytime and nighttime data. Hu et al. \cite{hu2020ddd20} built upon the DDD17 dataset \cite{binas2017ddd17} to create DDD20, which includes event-based VPR data collected in various locations in southwestern United States. It encompasses both daytime and nighttime data and spans 51 hours of footage, covering 4000km. Gehrig et al. \cite{gehrig2021dsec} introduced the DSEC dataset, an event-based stereo driving dataset situated in and around Zurich City, Switzerland. It consists of 53 sequences of daytime and nighttime event footage with a total duration of 0.887 hours. Verna et al. \cite{Verma_2024_CVPR} introduced e-TraM, a large-scale traffic monitoring event dataset that features 10 hours of data from both ego and static perspectives.


\section{NYC-Event-VPR}
\begin{figure}[t]
    \centering
    \includegraphics[width=\columnwidth]{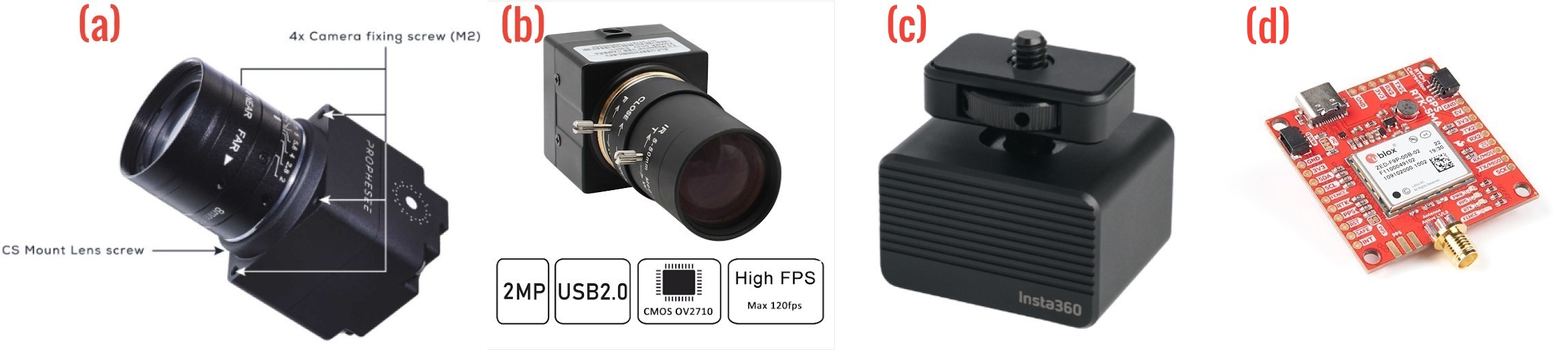}
    \caption{\textbf{Our sensor suite}: (a) Prophesee EVK4 HD Event Camera, (b) ELP USB Camera, (c) Insta360 vibration damper, (d) Sparkfun GPS-RTK-SMA GPS module}
    \label{fig:sensors}
\end{figure}

Our dataset is named NYC-Event-VPR. It encompasses large portions of Manhattan and Brooklyn in New York City, as illustrated in Fig. \ref{fig:coverage}. In comparison to other event-based VPR datasets mentioned on Sec. \ref{sec:related}, NYC-Event-VPR stands out in several aspects. We utilize the highest-resolution event camera with a resolution of 1280x720 pixels. Additionally, NYC-Event-VPR dataset includes repeated traversals spanning several months, encompassing both daytime and nighttime conditions, as well as varying weather scenarios (sunny, cloudy, rainy). NYC-Event-VPR's unique setting in downtown New York City introduces distinct features such as heavy traffic, pedestrian presence, and dense urban high-rise backgrounds, making it a valuable resource for event-based VPR research. Tab. \ref{table:statistics} provides an overview of NYC-Event-VPR's distinctive attributes. It encompasses more than 13.5 hours of multi-sensor data readings, incorporating data from forward-facing event camera and RGB camera, and a GPS module as shown in Fig. \ref{fig:sensors}.

\begin{table*}[t]
\caption{\textbf{NYC-Event-VPR dataset statistics}}
\vspace{-3mm}
\label{table:statistics}
\begin{center}
\begin{tabular}{c c c c c c c}
\toprule
\textbf{Duration (hr)} & \textbf{Data size (GB)} & \textbf{Modality} & \textbf{Distance (km)} & \textbf{Weather} & \textbf{Lighting conditions} & \textbf{Resolution (px)}\\
\hline
13.5 & 466.7 & event, RGB, GPS & 259.95 & rainy, cloudy, sunny & day, night & 1280 x 720 \\
\bottomrule
\end{tabular}
\end{center}
\vspace{-6mm}
\end{table*}

\subsection{Sensors}

\begin{figure}[t]
    \centering
    \includegraphics[width=\columnwidth]{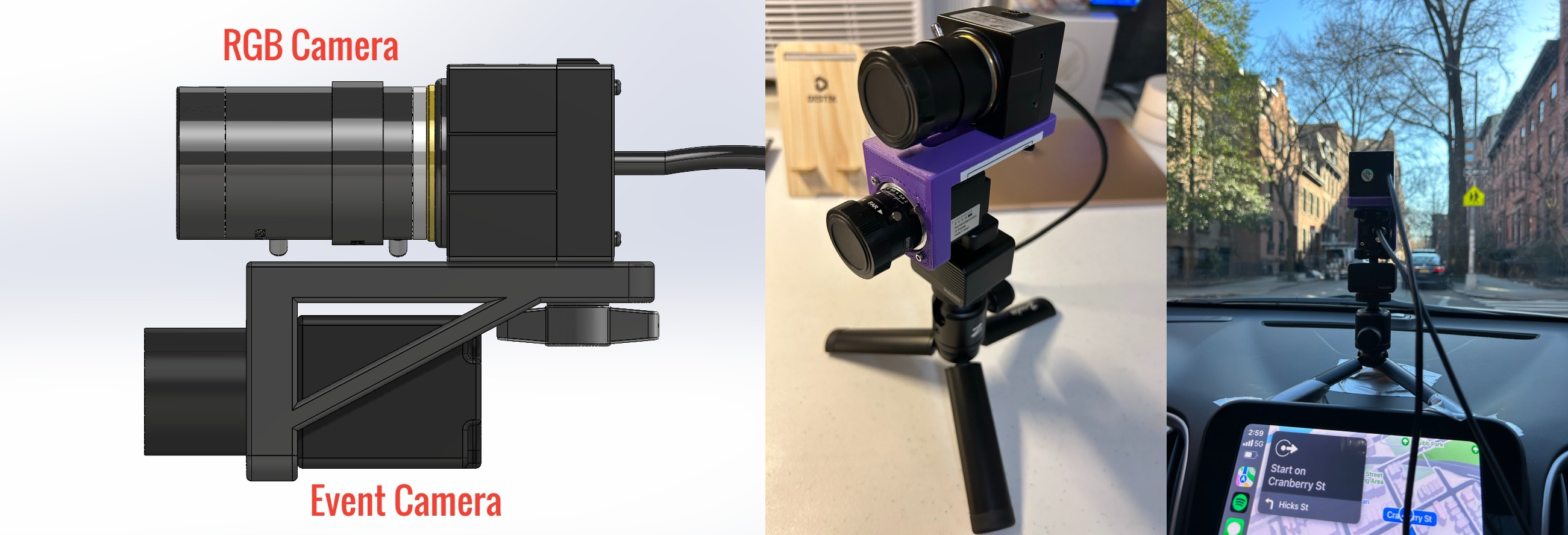}
    \caption{\textbf{Sensor setup and mounting design}: RGB camera is mounted on top of event camera, and the sensor suite is positioned facing forward behind vehicle's front windshield}
    \vspace{-5mm}
    \label{fig:setup}
\end{figure}

\begin{table}[t]
\caption{\textbf{Sensor specifications}}
\vspace{-3mm}
\label{table:sensor_specs}
\begin{center}
\begin{tabular}{ c c }
\toprule
\textbf{Type} & \textbf{Specification}\\
\hline
\multirow{8}{4em}{Prophesee EVK4 HD} &
    IMX636ES (HD) event vision sensor \\
    & Resolution (px): 1280x720 \\
    & Latency (\textmu s): 220 \\
    & Dynamic range (dB):  $>$86 \\
    & Power consumption: 500mW-1.5W \\
    & Pixel size (\textmu m): 4.86x4.86 \\
    & Camera max bandwidth (Gbps): 1.6 \\
    & Interface: USB 3.0
 \\
\hline
\multirow{4}{4em}{ELP USB camera} &
    CMOS 1080p sensor \\
    & Resolution (px): 1280x720 \\
    & Interface: USB 2.0 \\
    & 5-50mm varifocal lens 
\\
\hline
\multirow{4}{4em}{Sparkfun GPS-RTK-SMA} &
    Horizontal accuracy: 2.5m w/o RTK \\
    & Max altitude: 50km \\
    & Max velocity: 500m/s \\
    & GPS, GLONASS, Galileo, BeiDou
\\
\bottomrule
\end{tabular}
\end{center}
\vspace{-2mm}
\end{table}

\textbf{Event camera.} At the heart of our sensor suite setup is the event camera, specifically the IMX636 HD event sensor manufactured by Sony and Prophesee \cite{finateu20205}, part of the EVK4 HD evaluation kit, as detailed in Tab. \ref{table:sensor_specs}. Notably, unlike earlier event camera models with lower output resolutions, the IMX636 sensor impressively outputs event data at a resolution of 1280x720.

\textbf{RGB camera.} To seamlessly integrate the event camera into our setup, we design and 3D-print a custom structure as depicted in Fig. \ref{fig:setup}. Atop the event camera, we embed an ELP RGB camera that matches the event camera's resolution, also at 1280x720. To ensure precise alignment, we meticulously calibrate both cameras, synchronizing their focal lengths in pixel units, as illustrated in camera intrinsic matrices (1) and (2).

\vspace{-6mm}
$${\scriptstyle
K_{event} = \begin{bmatrix}
    1715 & 0 & 628 \\
    0 & 1715 & 375 \\
    0 & 0 & 1
\end{bmatrix}\hspace{1mm}{(1)} \hspace{3mm}
K_{rgb} = \begin{bmatrix}
    1998 & 0 & 639 \\
    0 & 1911 & 734 \\
    0 & 0 & 1
\end{bmatrix}\hspace{1mm}{(2)}} \vspace{5mm}
$$
\vspace{-7mm}

\textbf{Sensor Setup.} Beneath the sensor suite, we strategically mount an Insta360 vibration damper. This addition serves to mitigate data noise stemming from vehicle vibrations during travel, particularly in traffic conditions. The entire setup, including both cameras and the vibration damper, is securely affixed to a tripod as shown in Fig. \ref{fig:setup}. This arrangement ensures the stability and immovability of the setup within the data collection vehicle.

Positioning the sensor suite immediately behind the vehicle's front windshield facing forward, similar to the configuration in Brisbane-Event-VPR \cite{fischer2020event}, is a deliberate choice. This setup allows us to collect data even in adverse weather conditions such as rain, while avoiding potential water damage risks associated with a top-mounted configuration \cite{geiger2013vision}.

\textbf{GPS Module.} For geotagging purposes, we rely on Sparkfun's GPS-RTK-SMA breakout board in conjunction with a GNSS multi-band magnetic antenna. We opt not to enable Real-Time Kinematic (RTK) functionality since the board without RTK already offers meter-level accuracy, which suffices for our VPR objectives. The GPS board is housed inside the vehicle, while the antenna is securely affixed to the vehicle's roof, as shown in Fig. \ref{fig:setup}.

\textbf{Misc. Hardware.} To orchestrate the data collection process, we utilize a basic laptop, serving as a central hub for the three sensors. The sensors are connected to the hub laptop via USB 3.0 cables. To power this setup and ensure its operational efficiency, we employ a 300W portable power station, which provides the necessary power for both the laptop and the sensors.

\subsection{Data Collection}

\begin{table}[t]
\caption{\textbf{IMX636 event sensor bias settings}}
\vspace{-3mm}
\label{table:bias}
\begin{center}
\begin{tabular}{c c c c c c}
\toprule
\textbf{bias\_fo} & \textbf{bias\_hpf} & \textbf{bias\_diff\_off} & \textbf{bias\_diff\_on} & \textbf{bias\_refr} \\
\hline
-35 & 30 & 40 & 40 & 0 \\
\bottomrule
\end{tabular}
\end{center}
\vspace{-6mm}
\end{table}

\textbf{Collection Strategies.} Our approach to data collection for NYC-Event-VPR is thoughtfully designed to maximize scenery and condition diversity. We focus our efforts on Manhattan's vibrant Chinatown area, shown in Fig. \ref{fig:coverage}, ensuring overlapping coverage across multiple collection days to capture a wide array of scenarios and environments. Daytime and nighttime data collection further adds to the dataset's versatility, showcasing a broad range of lighting conditions. Typical sessions begin in the early afternoon and extend into dusk, sunset, and evening, comprehensively representing various lighting scenarios.

\textbf{Sensor Synchronization.} In real-time, the RGB camera captures intensity frames at a rate of 1Hz, synchronizing with the GPS module, also operating at 1Hz. In contrast, event data is recorded with only the starting timestamps for each event sequence since events inherently include timestamps with \textmu s precision. Post-processing techniques are then employed to synchronize the timestamps of event data with the GPS timestamps, ensuring accurate alignment and coordination between all the sensor modalities.

\textbf{Challenges.} One significant challenge we encounter during data collection is the substantial size of the output event data. In heavy Manhattan traffic, the event camera often generates events at a rate of up to 20 million events per second (MEv/s). To prevent bottlenecks in data transmission through USB 3.0 cables and alleviate CPU load, we implement several measures:

\begin{enumerate}
    \item Application of anti-flicker filters to eliminate signals between 50Hz and 500Hz.
    \item Implementation of event trail filters to remove trailing event bursts within a 100ms window.
    \item Utilization of an event rate controller to cap the maximum event rate at 10MEv/s.
    \item Fine-tuning hardware-level sensor biases (Tab. \ref{table:bias}).
    \item Adoption of concurrent programming techniques to distribute the computational workload across multiple CPU cores, enhancing overall efficiency.
\end{enumerate}

\section{Benchmark Experiments}

\subsection{Experiment setup}

\begin{figure}[t]
    \centering
    \includegraphics[width=1\columnwidth]{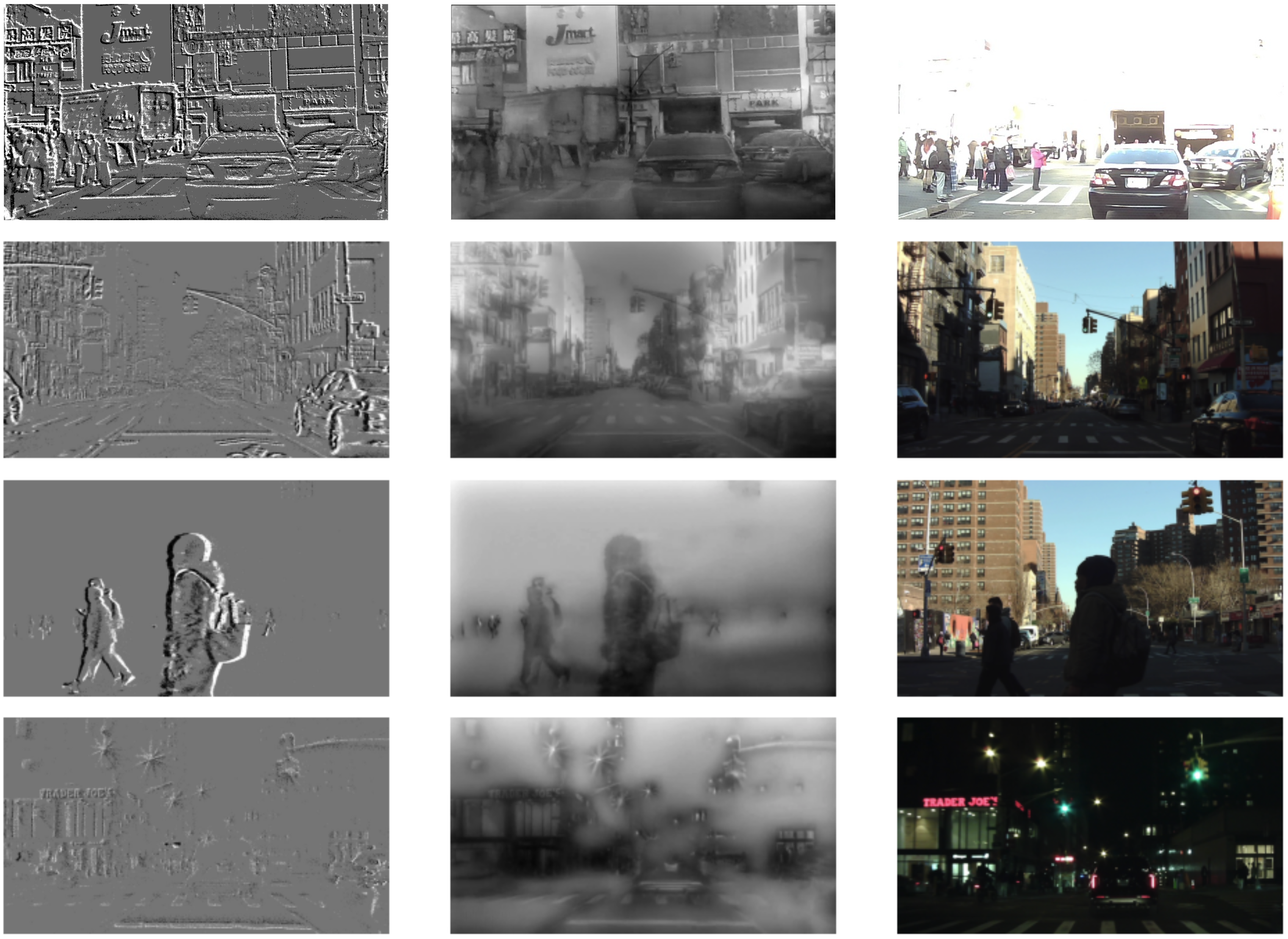}
    \caption{\textbf{Example images in processed dataset}~(from left to right columns): naive conversion, E2VID reconstruction, RGB reference. Each row is the same visual scene. Each column is the same dataset.}
    \vspace{-5mm}
    \label{fig:data_samples}
\end{figure}

\begin{table*}[h]
\caption{\textbf{Quantitative results of AUC-PR in percentage}. The best method for each dataset is in bold. Rows a-f are from NYC-Event-VPR. Rows g-h are curated subsets of their respective datasets provided by VPR-Bench. All tests are done using pretrained weights provided by VPR-Bench~\cite{zaffar2021vpr}.}
\label{table:results}
\centering
\NiceMatrixOptions{notes/para,notes/enumitem-keys-para={itemjoin = ;\;}}
\begin{NiceTabular}{c c c c c c c}
    \toprule
    \RowStyle{\bfseries}
    Datasets & NetVLAD~\cite{arandjelovic2016netvlad} & RegionVLAD~\cite{khaliq2019holistic} & HOG~\cite{dalal2005histograms} & AMOSNet~\cite{chen2017deep} & HybridNet~\cite{chen2017deep} & CALC~\cite{merrill2018lightweight} \\ \midrule
    NYC-Event-VPR-Naive-5m \tabularnote{naively rendered, sampled at 1fps, fault tolerance 5m} & 34.24 & 26.41 & 25.31 & 67.62 & 69.65 & \textbf{74.78} \\
    NYC-Event-VPR-Naive-15m \tabularnote{naively rendered, sampled at 1fps, fault tolerance 15m} & 40.53 & 29.84 & 32.14 & 73.09 & 75.09 & \textbf{81.55}\\
    NYC-Event-VPR-Naive-25m \tabularnote{naively rendered, sampled at 1fps, fault tolerance 25m} & 40.52 & 31.15 & 32.75 & 73.25 & 74.66 & \textbf{80.76}\\
    NYC-Event-VPR-E2VID-5m \tabularnote{E2VID reconstructed, sampled at 1fps, fault tolerance 5m} & 74.06 & 77.56 & 86.03 & \textbf{86.22} & 85.51 & 84.69\\
    NYC-Event-VPR-E2VID-15m \tabularnote{E2VID reconstructed, sampled at 1fps, fault tolerance 15m} & 84.92 & 89.15 & 95.89 & \textbf{97.16} & 96.87 & 96.09\\
    NYC-Event-VPR-E2VID-25m \tabularnote{E2VID reconstructed, sampled at 1fps, fault tolerance 25m} & 87.01 & 89.64 & 98.88 & \textbf{99.53} & 99.45 & 99.27\\
    NYC-EventVPR-RGB-5m \tabularnote{RGB reference, 1fps, fault tolerance 5m} & 92.52 & 92.58 & 92.43 & \textbf{94.52} & 94.33 & 92.86 \\
    NYC-EventVPR-RGB-15m \tabularnote{RGB reference, 1fps, fault tolerance 15m} & 98.14 & 97.12 & 95.20 & 98.26 & \textbf{98.26} & 96.85\\ 
    NYC-EventVPR-RGB-25m \tabularnote{RGB reference, 1fps, fault tolerance 25m} & 98.63 & 97.88 & 95.94 & \textbf{99.31} & 99.29 & 97.95 \\ \midrule
    Pittsburgh250K~\cite{arandjelovic2016netvlad} \tabularnote{Pittsburgh250K: RGB frames, 1,000 query images, 23,000 reference images} & \textbf{94.36} & 73.34 & 0.27 & 8.53 & 8.70 & 2.05 \\
    Nordland~\cite{Skrede_2013} \tabularnote{Nordland: RGB frames, 2,760 query images, 27,592 reference images} & 8.49 & 13.33 & 2.89 & \textbf{30.13} & 17.52 & 12.91 \\ \bottomrule
\end{NiceTabular}
\label{table:metrics}
\end{table*}

\begin{figure*}
    \centering
    \includegraphics[width=0.9\linewidth, height=3cm]{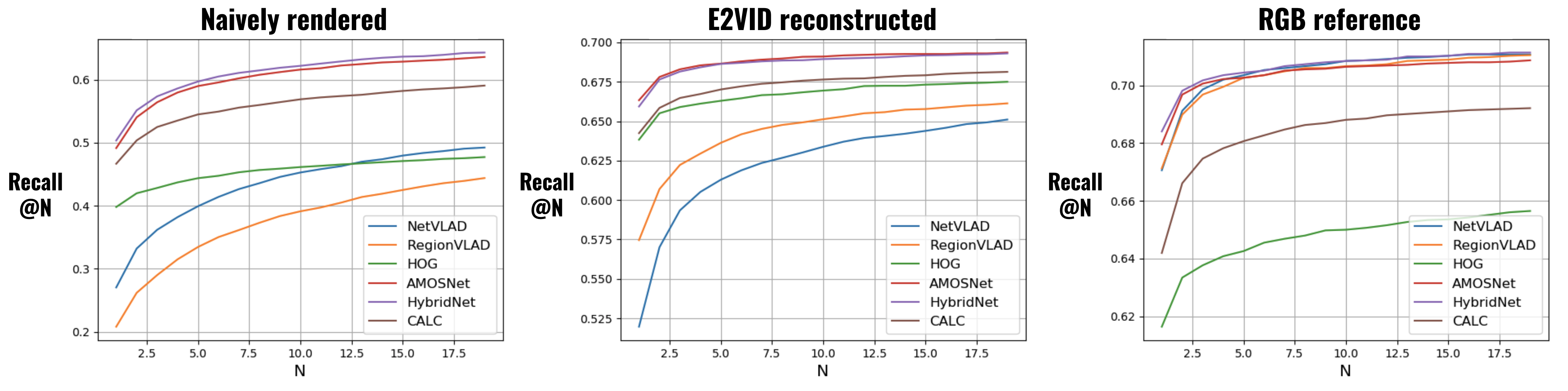}
    \caption{\textbf{Recall@N of VPR methods on naively rendered, E2VID \cite{rebecq2019high} reconstructed, and RGB reference datasets} with fault tolerance of 5m, sampled at 1fps.}
    \label{fig:recall}
\end{figure*}

\begin{table*}[h]
\caption{\textbf{Quantitative results of recall@N in percentage for CCT384\cite{cct2021} + NetVLAD\cite{arandjelovic2016netvlad} model and ResNet50\cite{he2016deep} + NetVLAD\cite{arandjelovic2016netvlad} model}. Each row represents each model's performances under 4 precision metrics. All benchmarks are done by training the model (backbone: CCT384 and ResNet50, aggregation: NETVLAD) on NYC-Event-VPR dataset using the Deep Visual Geo-localization Benchmark framework \cite{Berton_CVPR_2022_benchmark}.}
\label{table:trained_results}
\centering
\NiceMatrixOptions{notes/para,notes/enumitem-keys-para={itemjoin = ;\;}}
\begin{NiceTabular}{c c c c c c c c c}
    \toprule
    \RowStyle{\bfseries}
    & \multicolumn{4}{c}{\textbf{CCT384\cite{cct2021} + NetVLAD\cite{arandjelovic2016netvlad}}} & \multicolumn{4}{c}{\textbf{ResNet50\cite{he2016deep} + NetVLAD\cite{arandjelovic2016netvlad}}} \\
    \cmidrule(rl){2-5}  \cmidrule(rl){6-9}
    Datasets & Recall@1 & Recall@5 & Recall@10 & Recall@20 & Recall@1 & Recall@5 & Recall@10 & Recall@20\\ \midrule
    NYC-Event-VPR-Naive-5m & 33.2 & 45.1 & 48.7 & 51.2 & 39.1 & 48.9 & 51.5 & 53.1\\
    NYC-Event-VPR-Naive-15m & 51.0 & 64.0 & 68.6 & 73.2 & 54.9 & 68.6 & 72.6 & 75.8\\
    NYC-Event-VPR-Naive-25m & 57.3 & 74.1 &79.5 & 83.9 & 62.3 & 76.4 & 80.7 & 83.7\\
    NYC-Event-VPR-E2VID-5m & 48.7 & 54.4 & 55.8 & 57.1 & 51.9 & 54.7 & 55.6 & 56.2\\
    NYC-Event-VPR-E2VID-15m & 70.5 & 78.2 & 80.3 & 81.7 & 73.1 & 78.4 & 79.8 & 80.9\\
    NYC-Event-VPR-E2VID-25m & 77.9 & 85.4 & 87.1 & 88.7 & 80.5 & 85.8 & 87.5 & 89.0\\
    NYC-EventVPR-RGB-5m & 57.7 & 61.3 & 62.4 & 63.2 & 59.1 & 62.2 & 62.5 & 62.8\\
    NYC-EventVPR-RGB-15m & 77.8 & 81.4 & 82.4 & 83.1 & 79.0 & 82.4 & 83.1 & 83.5\\
    NYC-EventVPR-RGB-25m & 86.0 & 90.1 & 91.0 & 91.7 & 86.9 & 91.0 & 91.5 & 92.0\\ \bottomrule
\end{NiceTabular}
\label{table:trained_metrics}
\end{table*}

\begin{table*}[h]
\caption{\textbf{Quantitative results of recall@N in percentage for AnyLoc \cite{keetha2023anyloc} model}. Each row represents model performances under 4 precision metrics. All benchmarks are done by evaluating the model (backbone: DINOv2 \cite{oquab2023dinov2}) on NYC-Event-VPR dataset using the VPR Methods Evaluation framework \cite{Berton_2023_EigenPlaces}.}
\label{table:anyloc_results}
\centering
\NiceMatrixOptions{notes/para,notes/enumitem-keys-para={itemjoin = ;\;}}
\begin{NiceTabular}{c c c c c}
    \toprule
    \RowStyle{\bfseries}
    Datasets & Recall@1 & Recall@5 & Recall@10 & Recall@20 \\ \midrule
    NYC-Event-VPR-Naive-5m & 29.6 & 40.9 & 43.6 & 46.9 \\
    NYC-Event-VPR-Naive-15m & 45.8 & 60.4 & 64.6 & 68.7 \\
    NYC-Event-VPR-Naive-25m & 51.5 & 67.1 & 71.6 & 76.1 \\
    NYC-Event-VPR-E2VID-5m & 43.7 & 50.5 & 52.2 & 53.7 \\
    NYC-Event-VPR-E2VID-15m & 63.5 & 72.1 & 74.5 & 76.5 \\
    NYC-Event-VPR-E2VID-25m & 69.4 & 79.4 & 82.0 & 83.9 \\
    NYC-EventVPR-RGB-5m & 59.4 & 62.6 & 63.1 & 63.3 \\
    NYC-EventVPR-RGB-15m & 79.5 & 82.7 & 83.0 & 83.4 \\
    NYC-EventVPR-RGB-25m & 87.9 & 91.3 & 92.0 & 92.5 \\ \bottomrule
\end{NiceTabular}
\label{table:anyloc_metrics}
\vspace{-3mm}
\end{table*}

\begin{figure*}
    \centering
    \includegraphics[width=0.32\linewidth]{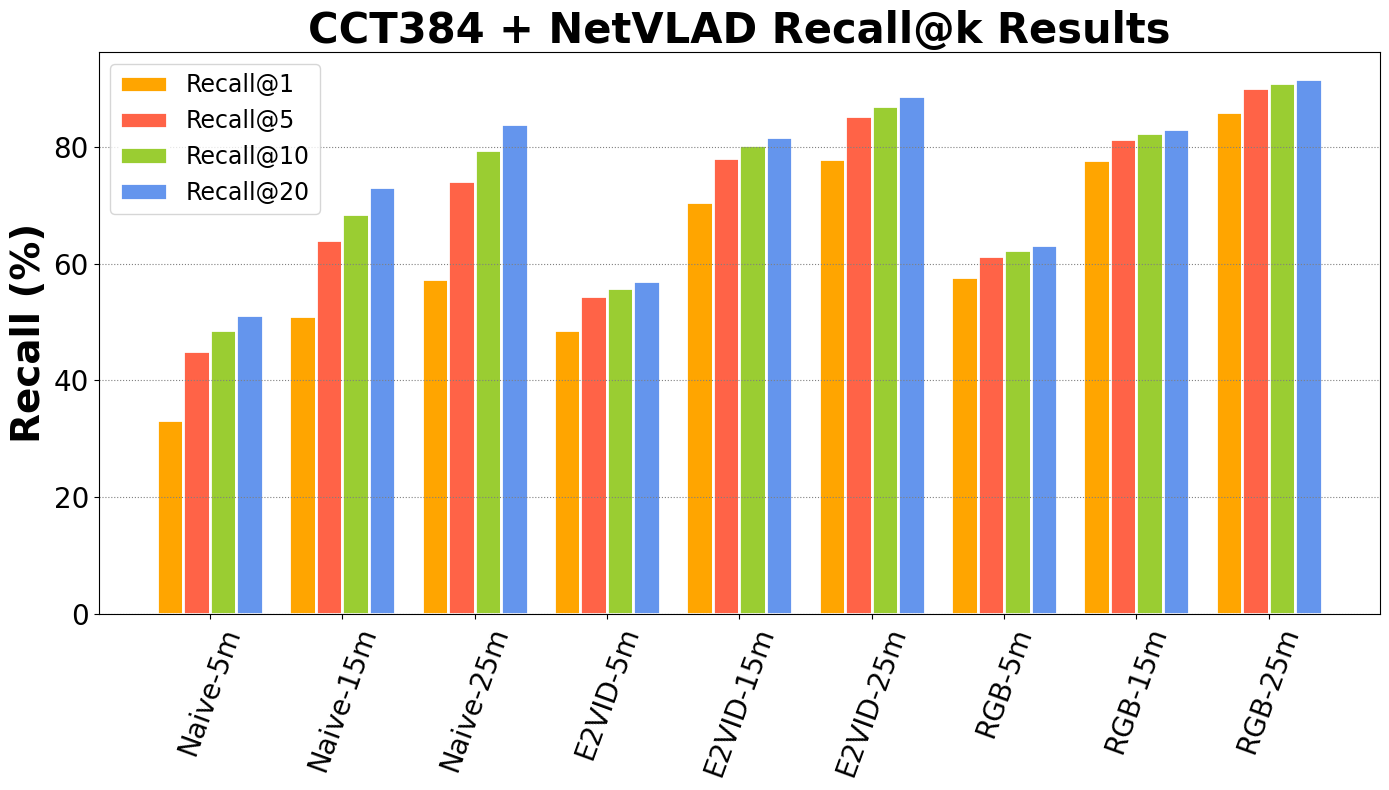}
    \includegraphics[width=0.32\linewidth]{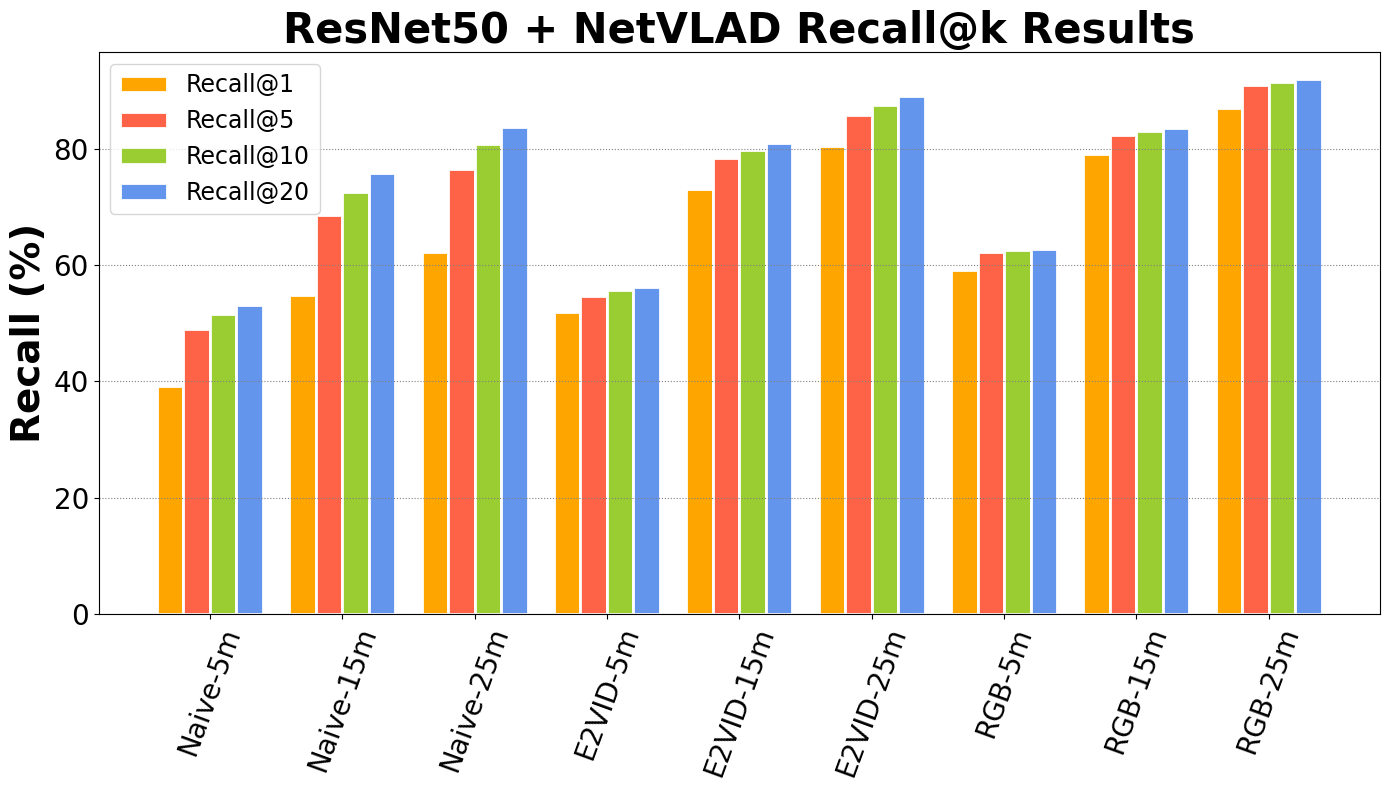}
    \includegraphics[width=0.32\linewidth]{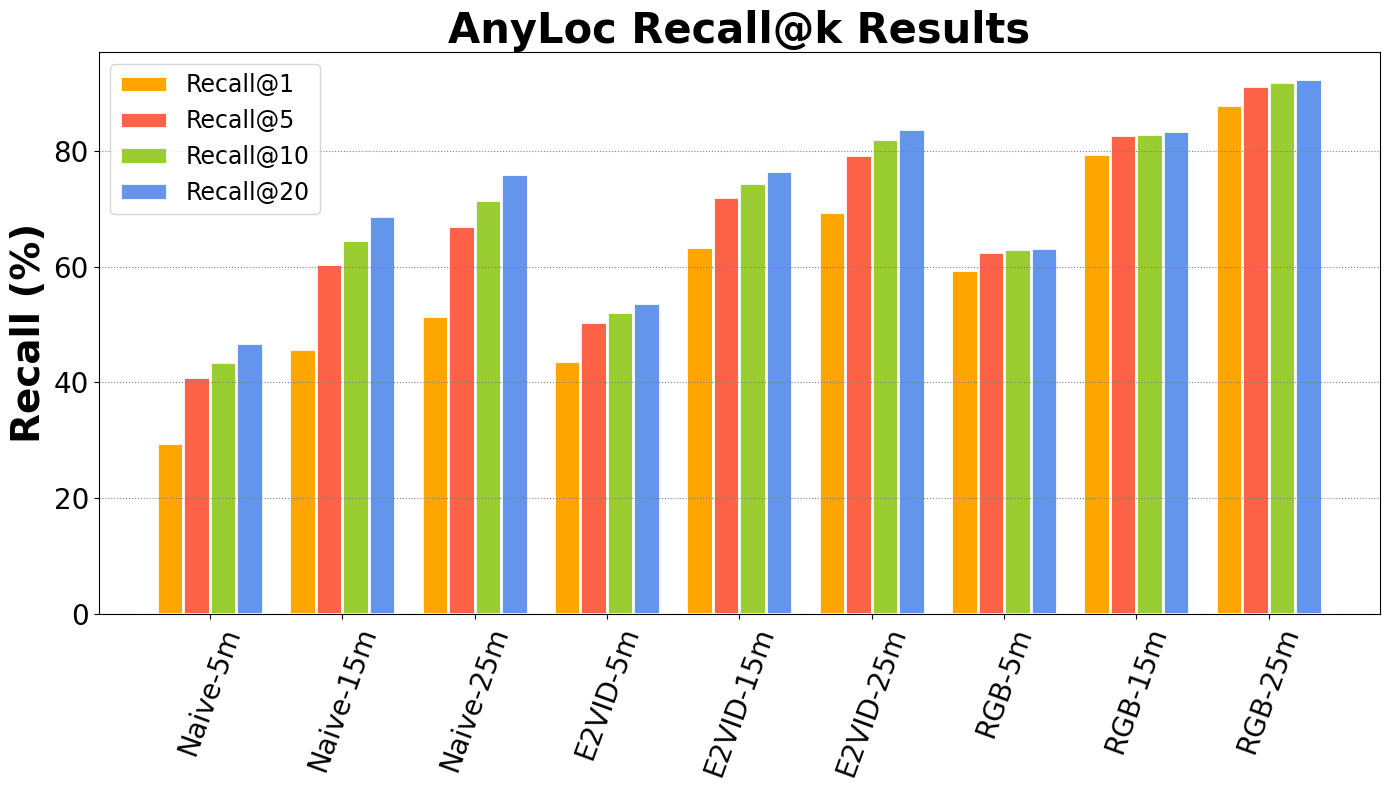}
    \caption{\textbf{Recall@N in percentage for CCT384\cite{cct2021} + NetVLAD\cite{arandjelovic2016netvlad},  ResNet50\cite{he2016deep} + NetVLAD\cite{arandjelovic2016netvlad}, and AnyLoc\cite{keetha2023anyloc} model}}
    \label{fig:plots}
    \vspace{-3mm}
\end{figure*}

\textbf{Hardware.} Our hardware setup for benchmarking the NYC-Event-VPR dataset consists of the following components: a Ryzen 9 7900X CPU, a RTX 4090 24GB GPU, 64GB of DDR5 RAM, and 4TB of M.2 NVMe SSD.

\textbf{Frameworks.} To facilitate preliminary benchmarking on our novel dataset, we leverage the VPR-Bench framework \cite{zaffar2021vpr}, the Deep Visual Geo-localization Benchmark \cite{Berton_CVPR_2022_benchmark}, and the VPR Methods Evaluation framework \cite{Berton_2023_EigenPlaces}. To make our dataset compatible with the frameworks, we pre-process it into a format that seamlessly integrates with the framework.

\textbf{VPR-Bench.} Developed by Zaffer et al., this framework offers the advantage of benchmarking our dataset using a wide array of traditional VPR techniques, including NetVLAD~\cite{arandjelovic2016netvlad}, RegionVLAD~\cite{khaliq2019holistic}, HOG~\cite{dalal2005histograms}, CoHOG \cite{zaffar2020cohog}, AlexNet \cite{krizhevsky2012imagenet}, AMOSNet~\cite{chen2017deep}, HybridNet~\cite{chen2017deep}, and CALC~\cite{merrill2018lightweight}. The weights of deep learning-based VPR methods in VPR-Bench are pretrained and frozen, and not trained on NYC-Event-VPR. 
\textbf{Deep Visual Geo-localization Benchmark.} Proposed by Berton et al., this framework supports experimentation on model performances with various choices of backbones and aggregation methods. Following their paper ~\cite{Berton_CVPR_2022_benchmark}, we train our data with CCT \cite{cct2021} and ResNet \cite{he2016deep} as backbone. NetVLAD \cite{arandjelovic2016netvlad} is selected as method of aggregation.

\textbf{VPR Methods Evaluation.} Introduced by Berton et al., this framework extends on the VPR-Bench framework, enabling the benchmarking of our dataset with the latest VPR techniques. In this study, we investigate the state-of-the-art AnyLoc model \cite{keetha2023anyloc}, a pre-trained foundation model.

\textbf{Intensity frames.} In line with the prevalent approach in event-based VPR, which often involves converting raw events into intensity frames and subsequently applying conventional frame-based VPR techniques~\cite{fischer2020event,lee2021eventvlad,rebecq2019high,scheerlinck2020fast}, we adopt a similar data processing pipeline. Note that VPR methods that operate directly on raw event streams \cite{hussaini2022spiking} also exist, potentially opening avenues for future research.

\textbf{Naive conversion.} For our initial baseline performance, we employ Prophesee's Metavision Studio software to perform straightforward conversions from raw events to intensity frames. In this baseline, frames are rendered in gray-scale at 30fps, implying that the software bins all events over a temporal window of 33.33ms and accumulates them into a single intensity frame.

\textbf{E2VID reconstruction.} Moving beyond the naive baseline, we utilize Prophesee's pretrained event-to-video model, an implementation of the E2VID model developed by Rebecq et al. \cite{rebecq2019high} This recurrent, fully convolutional network is trained on simulated event data and achieves state-of-the-art performance in conversion tasks from raw events to intensity frames. We apply both the naive conversion and the E2VID video reconstruction to the entirety of the raw event data from NYC-Event-VPR dataset. Fig. \ref{fig:data_samples} provides examples and a comparison with reference RGB frames.

Due to storage and computational limitations, our preliminary benchmarking experiments involve a sampling rate of 1fps for all dataset types: naively rendered, E2VID reconstructed, and RGB reference. A side-by-side comparison of three dataset types can be seen in Fig. \ref{figure:showcase}.




\textbf{Ground Truth.} To establish ground truth information and associate query and reference images, we utilize the GPS coordinates that are synchronized with timestamps and RGB images in real-time during data collection. We define a ``correct" match if the distance between query and reference images is below 25 meters, 15 meters, and 5 meters, considering these values as the fault tolerance thresholds. Given that event data are not synchronized in real-time during data collection, we perform timestamp interpolation to match naively rendered and E2VID reconstructed event frames with the temporally closest GPS-synchronized timestamps.


\subsection{Benchmark findings}

\textbf{Dataset Benchmark.} Benchmarking is performed on multiple datasets with varying fault tolerances, derived from NYC-Event-VPR raw data. Table~\ref{table:results} provides an overview of performance metrics measured in terms of Area Under the Curve - Precision Recall (AUC-PR) for various VPR methods. Fig. \ref{fig:recall} plots recall@N metrics of VPR methods used in benchmarking on naively rendered, E2VID reconstructed, and RGB datasets. Table~\ref{table:trained_results} presents the performance of the two trained models across four recall precisions. Table~\ref{table:anyloc_results} provides insights on the performance of AnyLoc across four recall precisions.Runs in Table~\ref{table:trained_results} and Table~\ref{table:anyloc_results} are evaluated on the same test set, with results visualized in Fig. \ref{fig:plots}. In all benchmark runs, a 10\% query and 90\% reference split is employed.  We summarize the following key insights from the benchmark experiments:

\begin{enumerate}
    \item \textbf{NYC-Event-VPR-RGB}: This dataset comprises reference images captured using an RGB camera. 
    Notably, under the same fault tolerance, models yield highest precisions when evaluated on RGB data. At a 5m fault tolerance, the NetVLAD method achieves an AUC-PR value of 34.24$\%$ when evaluated on naively rendered data, but it achieves an AUC-PR value of 92.52$\%$ when evaluated on RGB data. This outcome aligns with expectations as AnyLoc and the VPR methods in the VPR-Bench framework are designed for RGB data. 
    
    \item \textbf{NYC-Event-VPR-Naive}: In this dataset, event frames are generated through a straightforward conversion process involving temporal accumulation from raw event streams.  The lower precision results from event data align with our expectations, as the VPR methods used are originally designed for intensity frames. The naive conversion process from raw events to gray-scale intensity frames results in a substantial loss of temporal information, as precise timestamps (in microseconds) associated with each pixel-level event are discarded.
    \item \textbf{NYC-Event-VPR-E2VID}: The event dataset in this category is reconstructed from raw event streams using the E2VID recurrent convolutional model. This dataset demonstrates respectable AUC-PR performance metrics when assessed with conventional frame-based VPR methods. Specifically, the AMOSNet method achieves an AUC-PR metric of 86.22$\%$ at a 5m fault tolerance in this dataset. These findings are promising, as they indicate that the E2VID model, relying solely on raw events, is capable of reconstructing a frame-based dataset with commendable performance. This suggests the potential for further enhancement by applying event-based VPR methods directly to the event dataset.
\end{enumerate}




\section{Conclusion and Impact}

This paper introduces the NYC-Event-VPR dataset, a \textit{large-scale high-resolution} event-based visual place recognition (VPR) dataset collected in the \textit{dense urban environments} of New York City, featuring \textit{variable lighting conditions} and \textit{overlapping traversals}. The primary objective is to bridge the existing data gap between cutting-edge event-based visual sensors and VPR research. We introduce a flexible data processing pipeline for preparing and training event data, and demonstrate its potential through benchmarking on state-of-the-art VPR models. While the focus of this work is on VPR applications, the downstream utility of our dataset extends beyond VPR. Event data can be leveraged for various computer vision and robotics tasks, encompassing feature detection and tracking \cite{tedaldi2016feature, alzugaray2018asynchronous, gehrig2018asynchronous, gehrig_low-latency_2024}, depth and optical flow estimation \cite{bardow2016simultaneous, gallego2018unifying}, monocular and stereo 3D reconstruction \cite{rebecq2018emvs, zhou2018semi, zhou2021event}, pose estimation and SLAM \cite{mueggler2017event, vidal2018ultimate}, image reconstruction \cite{scheerlinck2020fast, rebecq2019high}, motion segmentation \cite{mitrokhin2019ev, stoffregen2019event}, and neuromorphic control \cite{vitale2021event, chen2020event}.

\section{Future Directions}
In future work, we plan to train the NYC-Event-VPR dataset on additional state-of-the-art VPR models to provide a more comprehensive benchmark evaluation. Furthermore, we aim to demonstrate the applicability of our dataset to the broader range of downstream applications mentioned earlier.

Meanwhile, we plan to expand the dataset's coverage, incorporate multi-modal sensors and event-specific VPR methods. This includes techniques that convert events into frames \cite{fischer2020event, rebecq2019high, scheerlinck2020fast} and those working directly on event streams \cite{hussaini2022spiking}. We will also explore sequence-based matching methods~\cite{garg2022seqmatchnet}. 

\addtolength{\textheight}{-1cm}   






\section*{Acknowledgment}

The authors would like to thank Ruixuan Zhang and Yinhong Qin for their contributions in data collection, Hanwen Zhao for hardware setup assistance, and Pratyaksh Rao for valuable comments.


{\small
\bibliographystyle{IEEEtranN}
\balance
\bibliography{reference}
}

\end{document}